\documentclass[conference]{IEEEtran}
\IEEEoverridecommandlockouts
\usepackage{cite}
\usepackage{amsmath,amssymb,amsfonts}
\usepackage{graphicx}
\usepackage{booktabs}
\usepackage{textcomp}
\usepackage{array}
\usepackage{tabularx}
\usepackage{url}

\begin{document}

\title{Beyond ROC-AUC: Operating-Point Performance Reporting for Biometric Verification}

\author{\IEEEauthorblockN{Ajan Ahmed and Masudul H. Imtiaz}
\IEEEauthorblockA{Department of Electrical and Computer Engineering\\
Clarkson University, Potsdam, NY, USA\\
Email: aahmed@clarkson.edu and mimtiaz@clarkson.edu}}

\maketitle

\begin{abstract}
A biometric verification system is often deployed with a strict false match budget, so only a narrow, low false match rate (FMR) slice of the score range is used. A reporting standard for this setting already exists. ISO/IEC 19795-1 asks for error rates at stated operating points, for the detection error tradeoff (DET) curve as the view of the trade-off between FMR and the false non-match rate (FNMR), and for an interval of uncertainty on every value. In practice, a single area under the receiver operating characteristic curve (ROC-AUC), the equal error rate (EER), or a verification accuracy is still reported as the resolution, which is a threshold-independent summary that the standard does not endorse. The full ROC-AUC averages the true match rate (TMR) with equal weight over the whole FMR range from $0$ to $1$, so almost all of its weight is placed where the system is never operated; low-FMR behavior can then be hidden, and the order of two systems can even be reversed. The guideline is revisited in this paper and tested against seven pretrained matchers across four modalities, face, voice, iris, and fingerprint, each reported with bootstrap confidence intervals and paired bootstrap tests. A system that looks stronger on full ROC-AUC is shown to be significantly worse at $\mathrm{FMR}=10^{-3}$. For face, a higher full AUC was obtained by FaceNet, whereas a higher TMR at $\mathrm{FMR}=10^{-3}$ was obtained by ArcFace, and both gaps were significant with non-overlapping intervals. Hence, the DET curve and the FNMR at a fixed FMR are re-iterated in this paper as the primary report, with ROC-AUC and EER retained as supplementary context.
\end{abstract}

\begin{IEEEkeywords}
biometric verification, performance reporting, ROC-AUC, detection error tradeoff, operating point, ISO/IEC 19795
\end{IEEEkeywords}

\section{Introduction}

Biometric recognition verifies a person's identity from a physical or behavioral trait, such as a face image, a voice sample, an iris image, or a fingerprint. These systems now support applications such as border control, device unlocking, and account access~\cite{jain2004biometrics,ali2026fpga,chowdhury2022contactless,ketola2022eeg}. In one-to-one verification, two samples are compared, a similarity score is produced, and the claim is accepted when the score passes a threshold. Moving this threshold changes the balance between two errors. The false match rate (FMR) is the fraction of non-mated comparisons that are accepted, while the false non-match rate (FNMR) is the fraction of mated comparisons that are rejected. The true match rate (TMR) is $1-\mathrm{FNMR}$. These terms follow the international standard for biometric performance testing and reporting~\cite{isoiec19795}.

A reporting guideline gives researchers and practitioners a common way to describe system performance, compare matchers, and plan deployment thresholds. In biometrics, this guidance already exists. ISO/IEC 19795-1 asks reports to include error rates at stated operating points, the curve that relates FMR and FNMR, and an uncertainty interval for each reported value~\cite{isoiec19795}. In practice, however, many verification publications still use one resolution scalar, most often the area under the ROC curve (AUC)~\cite{fawcett2006roc,hanley1982meaning}, sometimes together with the equal error rate (EER), where FMR and FNMR are equal. The full AUC is useful as a threshold-independent summary because it averages TMR over all FMR values from $0$ to $1$. Yet this convenience also limits its usefulness for deployment, since the standard is concerned with the operating point rather than with the full score range.

This ambiguity can also have practical consequences. Table~\ref{tab:realworld_cases} summarizes representative settings in which a broad AUC statement, an unclear threshold, or an unreported low-FMR operating point can lead to a different operational reading of a biometric system.
\begin{table*}[t]
\centering
\caption{Representative real-world settings where AUC and FAR-threshold ambiguity can affect biometric decisions.}
\label{tab:realworld_cases}
\footnotesize
\setlength{\tabcolsep}{3pt}
\renewcommand{\arraystretch}{1.12}
\begin{tabularx}{\textwidth}{p{0.12\textwidth}p{0.18\textwidth}X X p{0.16\textwidth}}
\toprule
\textbf{Modality} & \textbf{Example setting} & \textbf{Metric or threshold problem} & \textbf{Possible application issue} & \textbf{Example references} \\
\midrule
Fingerprint & Latent-print or watch-list search followed by human review & Reporting only a global accuracy/AUC value can hide the tail of non-mated scores; a permissive candidate threshold may create plausible but incorrect leads. & A false candidate can be elevated from an investigative lead to an identification decision if the low-FAR operating point and uncertainty are not clearly reported. & Mayfield fingerprint misidentification review~\cite{doj2006mayfield}; fingerprint recognition overview~\cite{maltoni2009handbook} \\
Face & Law-enforcement face search or identity screening & A high full ROC-AUC may coexist with unacceptable false-match behavior at the deployment threshold, especially when searches are scaled to large galleries. & An erroneous face-recognition lead can contribute to wrongful arrest when treated as stronger evidence than the tested threshold supports. & Williams facial-recognition false-arrest case~\cite{aclu2024williams}; NIST FRTE reporting definitions~\cite{nistfrte} \\
Voice & Telephone-banking or remote speaker verification & Speaker verification can be vulnerable to mimicry, replay, synthesis, or conversion; full AUC on benign trials may not describe spoofing or stringent FAR operation. & A threshold selected from the wrong region can increase unauthorized access risk or produce excessive false rejects for legitimate users. & HSBC Voice ID bypass report~\cite{guardian2017hsbc}; ASVspoof 2019 database and protocols~\cite{wang2020asvspoof2019,ahmad2026deepfake,vijaykumar2026elad} \\
Iris & Border-control or high-assurance access screening & A strong decidability index or global curve summary may not show how capture quality and threshold choice affect FNMR and FMR in the deployed low-FMR region. & A threshold set outside the tested operating region can increase manual review and false rejects for legitimate users, while rare false matches can still matter at large scale. & Iris recognition operating principles~\cite{daugman2004iris}; iris biometrics survey~\cite{bowyer2008iris} \\
\bottomrule
\end{tabularx}
\end{table*}
The same care is needed when naming area-style metrics. Table~\ref{tab:metric_taxonomy} keeps the main AUC variants separate by integration interval, integration variable, normalization, and preferred direction.

\begin{table*}[t]
\centering
\caption{AUC-style summaries for biometric verification. Let $T(u)$ denote the interpolated TAR/TMR at FAR/FMR $u$. The same ROC points can produce different values depending on the interval, integration variable, and normalization.}
\label{tab:metric_taxonomy}
\footnotesize
\begin{tabular}{@{}llllp{6.2cm}@{}}
\toprule
Metric & Formula sketch & FAR interval & Direction & Main interpretation \\
\midrule
Full ROC-AUC & $\int_0^1 T(u)\,du$ & $[0,1]$ & high is better & General ranking summary; dominated by the wide high-FAR portion of the unit interval. \\
Raw pAUC & $\int_a^b T(u)\,du$ & user-selected $[a,b]$ & high is better & Region-specific area; its magnitude changes with interval width. \\
Normalized pAUC & $(b-a)^{-1}\int_a^b T(u)\,du$ & user-selected $[a,b]$ & high is better & Average TAR over a chosen linear-FAR interval. \\
Standardized pAUC & McClish-style scaling & usually $[0,b]$ & high is better & Software convention in which random performance is near 0.5 and perfect performance is 1. \\
Log-FAR AUC & $\Delta_{\log}^{-1}\int_a^b T(u)\,d\log_{10}u$ & $0<a<b\le 1$ & high is better & Average TAR per decade of FAR; emphasizes low-FAR behavior. \\
DET-style log error area & $\Delta_{\log}^{-1}\int_a^b [1-T(u)]\,d\log_{10}u$ & $0<a<b\le 1$ & low is better & Average FNMR per decade of FAR; error-oriented companion to log-FAR AUC. \\
Fixed-FAR point & $T(\alpha)$ or $1-T(\alpha)$ & single $\alpha$ & rate-dependent & Operational reporting at the deployment-relevant FMR/FAR. \\
\bottomrule
\end{tabular}
\end{table*}

Current reporting practice is also uneven across biometric modalities. Table~\ref{tab:metriczoo} lists representative verification studies and the metric each uses as its headline result. Face-related papers often report verification accuracy or a true accept rate at a fixed false accept rate; speaker verification papers often report EER and a detection cost; iris papers commonly report a decidability index; and fingerprint papers report EER together with fixed-FMR operating points. Because these quantities differ from study to study, and because confidence intervals are often omitted, it can be difficult to compare systems on the same practical basis.

\begin{table}[t]
\centering
\caption{Representative biometric verification studies and the headline metric used by each. No single quantity is shared across modalities, and most summarise the whole curve rather than the strict operating region, usually without an interval.}
\label{tab:metriczoo}
\footnotesize
\begin{tabular}{|p{2.0cm}|p{1.2cm}|p{3.3cm}|}
\hline
Study & Modality & Headline metric(s) \\
\hline
FaceNet~\cite{schroff2015facenet} & Face & Accuracy; TAR at FMR$=10^{-3}$ \\
\hline
ArcFace~\cite{deng2019arcface} & Face & Accuracy; TAR at FMR$=10^{-4}$--$10^{-6}$ \\
\hline
FRVT~\cite{grother2019frvt} & Face & FNMR at fixed FMR ($10^{-4}$--$10^{-6}$) \\
\hline
i-vector~\cite{dehak2011ivector} & Voice & EER; minDCF \\
\hline
x-vector~\cite{snyder2018xvector} & Voice & EER; minDCF \\
\hline
ECAPA-TDNN~\cite{desplanques2020ecapa} & Voice & EER; minDCF \\
\hline
VoxCeleb~\cite{nagrani2017voxceleb} & Voice & EER \\
\hline
NIST SRE~\cite{nist2018sre} & Voice & minDCF; actual DCF \\
\hline
Br\"ummer \& du Preez~\cite{brummer2006cllr} & Voice & $C_{\text{llr}}$ (calibration cost) \\
\hline
Daugman~\cite{daugman1993iris} & Iris & Decidability $d'$; Hamming distance \\
\hline
Daugman~\cite{daugman2004iris} & Iris & Decidability $d'$; FAR/FRR crossover \\
\hline
FVC2002~\cite{maio2002fvc2002} & Finger & EER; FMR100, FMR1000; ZeroFMR \\
\hline
Handbook~\cite{maltoni2009handbook} & Finger & EER; FMR/FNMR curves \\
\hline
Jain et al.~\cite{jain2004biometrics} & General & FAR/FRR; EER; ROC \\
\hline
\end{tabular}
\end{table}

\begin{figure}[tbp]
\centering
\includegraphics[width=\linewidth]{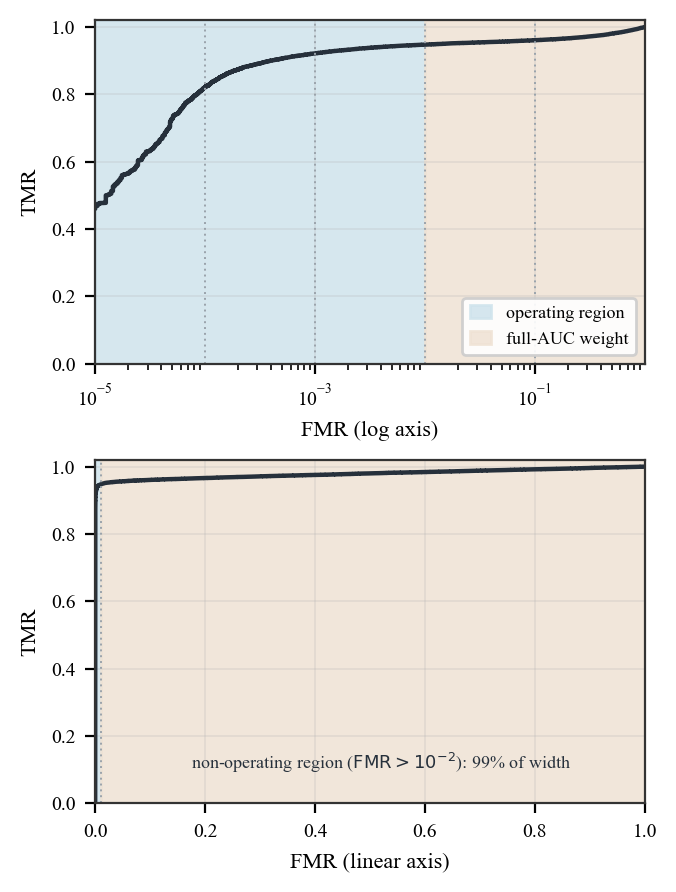}
\caption{The same face system (ArcFace on LFW) under two FMR axes. Top, a logarithmic axis, on which deployment is performed in the shaded operating region and the curve detail is visible. Bottom, a linear axis, on which the operating region ($\mathrm{FMR}\le10^{-2}$) is collapsed to a thin sliver and about $99$ percent of the width is occupied by the non-operating region. The weight of a full ROC-AUC is spread across this entire linear range, most of which is never used by the system.}
\label{fig:motivation}
\end{figure}

This mismatch matters because biometric systems are rarely operated over the full FMR range. Many deployments use a strict false match budget, often near $10^{-2}$, $10^{-3}$, or lower, because frequent false matches are unacceptable in settings such as account login or e-gate screening. On a linear FMR axis, the interval from $10^{-3}$ to $1$ covers $99.9$ percent of the width, so most of the full AUC is determined by regions where the system would not normally be used. Figure~\ref{fig:motivation} illustrates this compression. A system can therefore look strong under full AUC while performing poorly at the threshold that matters. In some cases, two systems can even be ranked in the opposite order. In the face experiments in this paper, the full AUC favors one matcher, while $\mathrm{TMR}$ at $\mathrm{FMR}=10^{-3}$ favors the other. The choice of headline metric can therefore affect the practical conclusion of a comparison.

For this reason, the evaluation should focus on the region in which the system will be used, and the uncertainty of each estimate should be made explicit. This is also the direction already taken by ISO/IEC 19795-1. In this paper, the DET curve, which shows FNMR against FMR in a way that makes the low-FMR region readable, is treated as the main graphical report. The main tabulated quantity is FNMR at a fixed FMR, with low-FMR partial AUC~\cite{mcclish1989portion}, log-FMR AUC, and minimum detection cost~\cite{nist2018sre} used as supporting summaries. Bootstrap confidence intervals are used to show how reliable each estimate is for the available number of trials~\cite{efron1993bootstrap}. The full ROC-AUC and EER are not rejected; instead, they are kept as supplementary context, consistent with a report that emphasizes the trade-off and its uncertainty rather than a single global scalar~\cite{guo2017calibration,gal2016dropout}.

This paper makes three contributions:
\begin{itemize}
\item it revisits ISO/IEC 19795-1 and contrasts the standard's operating-point focus with the metrics commonly used in verification papers;
\item it evaluates seven pretrained matchers across face, voice, iris, and fingerprint, reporting confidence intervals at the operating points and paired bootstrap tests between matchers;
\item it shows a statistically significant face-ranking flip, where the preferred system changes with the metric used as the headline result, and provides a short reporting checklist aligned with the standard.
\end{itemize}

\section{The Reporting Guideline}

ISO/IEC 19795-1 provides the principles and framework for biometric performance testing and reporting~\cite{isoiec19795}. Its purpose is not to promote one convenient score, but to make biometric results comparable by stating how a test was run, which operating point was used, and how much uncertainty is attached to the estimate.
The standard centers the report on error rates. At the comparison level, these are FMR and FNMR. At the system level, false accept and false reject rates are also reported, including effects such as failure to acquire and failure to enroll. These failure rates are kept visible rather than absorbed into the accept and reject rates. Performance is viewed as a trade-off between error rates as the decision threshold moves. For that reason, the natural report is a curve rather than a single point. The detection error tradeoff curve, which plots FNMR against FMR on a transformed scale, is important because the operating points of interest often lie at low FMR, where a linear plot is hard to interpret~\cite{martin1997det}.

The standard also asks that each reported number be tied to clear test conditions. The operating point should be stated, the statistical uncertainty should be given, and the dataset, population, protocol, number of subjects, and number of comparisons should be documented alongside the result. These details are important because low-FMR estimates depend strongly on the number of available non-mated trials.

A full ROC-AUC, a plain verification accuracy, or a single EER does not answer this deployment-focused question by itself. Such quantities summarize either the whole curve or a point that may not match the intended threshold. Nevertheless, matcher-level ROC-AUC is still often used as the headline result. The rest of this work examines this gap between recommended reporting and common practice, and shows how large the gap can be in practical comparisons.

\section{Related Work}

ROC analysis and AUC are well established for binary scores~\cite{fawcett2006roc}. The AUC also has a useful probabilistic interpretation: it is the probability that a randomly selected mated pair receives a higher score than a randomly selected non-mated pair~\cite{hanley1982meaning}. Its limitation in biometric verification is that the full area assigns equal weight to all operating points, while deployments usually care about only the strict end of the curve. Partial AUC focuses on a chosen false-positive range, and the standardized form rescales the area so that random performance lies near $0.5$~\cite{mcclish1989portion}. Speaker recognition also uses detection cost functions and calibration-aware costs to summarize performance at operating points~\cite{nist2018sre,brummer2006cllr,ahmed2026speechquality,ahmed2024vpqad,khondkar2026speech}. Iris recognition has often used the decidability index to describe the separation between mated and non-mated score distributions~\cite{daugman2004iris,daugman1993iris}. Related concerns appear in information retrieval and virtual screening, where early-recognition summaries such as concentrated ROC and Boltzmann-weighted ROC have been proposed~\cite{swamidass2010croc,truchon2007bedroc}. Class imbalance motivates another related line of work: the precision-recall curve, summarized by PR-AUC, can be more sensitive than the ROC curve when the negative class is much larger than the positive class~\cite{davis2006prroc,saito2015prroc}.

These tools are available, but they are not always reflected in the way verification results are reported. A full AUC may still appear without qualification, making it hard to read system behavior at the deployment threshold. Conversely, when a low-FMR number is reported, it is often presented without an interval, so the reader cannot tell whether a small difference is meaningful. Hence, we emphasize strict-threshold quantities as the main report, attach a bootstrap confidence interval to every value, and use a paired bootstrap test to compare systems at the operating point. Table~\ref{tab:contrast} summarizes the difference between the common style of reporting and the operating-point-centered report used here.

\begin{table}[t]
\centering
\caption{Common practice against the reporting used in this work.}
\label{tab:contrast}
\footnotesize
\begin{tabular}{@{}p{2.4cm}p{2.4cm}p{2.55cm}@{}}
\toprule
Aspect & Common practice & This work \\
\midrule
Primary report & Full ROC-AUC, EER scalar & DET curve, FNMR at fixed FMR \\
Role of ROC-AUC & Headline & Supplementary context \\
Region weighted & Whole FMR $[0,1]$ & Operating FMR $\le 10^{-2}$ \\
Uncertainty & Usually absent & Bootstrap CI on every value \\
System comparison & AUC ordering & Paired test at the threshold \\
Imbalance & Not addressed & ROC-AUC versus PR-AUC gap \\
\bottomrule
\end{tabular}
\end{table}

\section{Methodology}

\subsection{Datasets}
One public dataset was selected for each modality in this work so that the study could focus on reporting behavior rather than on a private benchmark. For face, we used Labeled Faces in the Wild (LFW), a standard unconstrained verification benchmark~\cite{huang2007lfw}. For voice, we used VoxCeleb1, a standard speaker verification dataset~\cite{nagrani2017voxceleb}. For iris, we used CASIA-Iris-Thousand from version four of the iris collection released by the Institute of Automation, Chinese Academy of Sciences (CASIA)~\cite{casia2010irisv4}; this collection is also included among the iris resources surveyed in~\cite{bowyer2008iris}. For fingerprint, we used the Set B images from the second Fingerprint Verification Competition (FVC2002), which provides several impressions per finger~\cite{maio2002fvc2002,maltoni2009handbook}. FVC2002 was chosen because repeated captures of the same finger are needed to form mated pairs for a classical minutiae matcher.

\subsection{Recognition systems}
We evaluated widely used pretrained matchers without additional training, so the comparison reflects openly available systems rather than a private training setup. For face, we used ArcFace~\cite{deng2019arcface} and FaceNet~\cite{schroff2015facenet}, with face detection and alignment performed by a multitask cascaded network~\cite{zhang2016mtcnn}. For voice, we used ECAPA-TDNN~\cite{desplanques2020ecapa} and the x-vector~\cite{snyder2018xvector} from the SpeechBrain toolkit~\cite{ravanelli2021speechbrain}, with the i-vector noted as their predecessor~\cite{dehak2011ivector}. For iris, we used the open-source pipeline released by the Worldcoin project~\cite{worldcoin2024openiris} and a classical log-Gabor encoder with Hamming distance in the style of Daugman~\cite{daugman1993iris}. For fingerprint, we used the open SourceAFIS minutiae matcher~\cite{vazan2023sourceafis}. The two-matchers-per-modality design provides a stronger and weaker system for the ranking comparison in face, voice, and iris. Fingerprint includes only one matcher because no widely used open deep matcher was available, so it supports within-system metric comparisons but not a ranking-flip test.

\subsection{Performance metrics}
Let $T(u)$ denote the interpolated TMR at FMR $u$. The selected metrics separate several ideas that are often combined in a single headline number. The full ROC-AUC is the average TMR over the full linear FMR range,
\begin{equation}
A_{\text{full}} = \int_0^1 T(u)\, du .
\end{equation}
The partial AUC over $[0,b]$ can be reported in different forms. Here, the normalised partial AUC (pAUC) is the partial area divided by its width, which is the average TMR below FMR $b$,
\begin{equation}
\bar{T}_{b} = \frac{1}{b}\int_0^{b} T(u)\, du ,
\end{equation}
rather than the standardised pAUC of McClish~\cite{mcclish1989portion}, which rescales the area so that random performance is mapped near $0.5$. We evaluate $\bar{T}_b$ at $b=10^{-1}$, $10^{-2}$, and $10^{-3}$. The log-FMR AUC integrates with respect to $\log_{10}\mathrm{FMR}$ and therefore gives equal weight to each decade of FMR,
\begin{equation}
A_{\log} = \frac{1}{\log_{10} b - \log_{10} a}\int_a^b T(u)\, d\log_{10} u ,
\end{equation}
with $a=10^{-3}$ and $b=1$, so the metric averages over three FMR decades.
The minimum detection cost (minDCF) is the cost-weighted operating point commonly used in speaker recognition~\cite{nist2018sre},
\begin{equation}
\text{minDCF} = \min_\tau \frac{C_{\text{m}}\,\mathrm{FNMR}(\tau)\,P_{\text{t}} + C_{\text{fa}}\,\mathrm{FMR}(\tau)\,(1-P_{\text{t}})}{\min(C_{\text{m}} P_{\text{t}},\, C_{\text{fa}}(1-P_{\text{t}}))},
\end{equation}
with target prior $P_{\text{t}}=0.01$ and unit costs. We evaluate TMR and FNMR at fixed FMR targets from $10^{-1}$ through $10^{-4}$. The recommended tabulated quantity is FNMR at a fixed FMR; the TMR reported here is its complement, $\mathrm{TMR}=1-\mathrm{FNMR}$. We also evaluate EER, decidability $d'$ for the separation of the two score distributions~\cite{daugman2004iris}, and PR-AUC as an imbalance-sensitive summary~\cite{davis2006prroc,saito2015prroc}. Two qualifications are important. PR-AUC depends on the ratio of mated to non-mated trials, so unlike ROC-AUC, it is not prevalence-free and should be interpreted together with that ratio~\cite{davis2006prroc,saito2015prroc}. minDCF depends on the chosen prior and cost pair, so it summarizes an operating point without necessarily demonstrating calibration~\cite{nist2018sre,brummer2006cllr}.

\subsection{System Verification}
For each system, we extracted one embedding or template per sample and then formed mated and non-mated pairs from identity labels. A mated pair contains two samples from the same identity, while a non-mated pair contains samples from different identities. Cosine similarity was used for the embedding systems, Hamming similarity for iris, and the matcher score for fingerprint. Scores were oriented so that larger values indicated stronger evidence for a mated pair. Large non-mated sets were sampled: two million pairs for face, one million for voice, and about $150{,}000$ for iris. For fingerprint, the dataset limited the protocol to about $14{,}000$ within-sensor non-mated pairs. 

For each face and voice modality, the same pair list was scored by both matchers, which keeps the system comparisons paired trial by trial. Before scoring, the pair pools contained $1{,}680$ identities over $9{,}164$ LFW images for face, $40$ speakers over $4{,}708$ VoxCeleb1 utterances for voice, $818$ eyes over $2{,}454$ CASIA images for iris, and $50$ fingers over $400$ FVC2002 images with eight impressions per finger. Non-mated pairs were drawn with a fixed seed, and the pair list was generated once rather than resampled before scoring. Face mated pairs were capped at $2{,}000$ per identity, and fingerprint non-mated pairs were restricted to the same database under a within-sensor protocol. Table~\ref{tab:proto} lists the final pair counts.

\begin{table}[t]
\centering
\caption{Verification protocol. Counts are the mated and non-mated pairs scored per matcher.}
\label{tab:proto}
\footnotesize
\begin{tabular}{@{}llrr@{}}
\toprule
Modality & Dataset & Mated & Non-mated \\
\midrule
Face & LFW & $59{,}489$ & $2{,}000{,}000$ \\
Voice & VoxCeleb1 & $310{,}095$ & $1{,}000{,}000$ \\
Iris & CASIA-Iris-Thousand & $2{,}454$ & $150{,}000$ \\
Fingerprint & FVC2002 Set B & $1{,}400$ & $14{,}400$ \\
\bottomrule
\end{tabular}
\end{table}

\subsection{Comparison and significance design}
We performed five comparisons. First, to measure overstatement, we placed the full AUC and the low-FMR averages on the same average-TMR scale and computed
\begin{equation}
G_b = A_{\text{full}} - \bar{T}_b ,
\end{equation}
which measures how far the full AUC lies above the strict-threshold value. Second, to study predictiveness, we identified which scalar best tracked the fixed-FMR operating points. Third, to study ranking, we checked whether the two matchers within a modality were ordered the same way by full AUC and by strict-threshold metrics. Fourth, to study imbalance, we measured the gap between ROC-AUC and PR-AUC. Fifth, to study uncertainty, we attached a confidence interval to every reported value. We used the percentile bootstrap for uncertainty estimation~\cite{efron1993bootstrap}. For each metric, $B=200$ resamples were drawn with replacement from the trial pairs, the metric $m^{(b)}$ was recomputed, and the interval was obtained as
\begin{equation}
\mathrm{CI}_{95}(m) = \left[\, Q_{2.5}\{m^{(b)}\},\; Q_{97.5}\{m^{(b)}\}\,\right].
\end{equation}
For comparisons between two matchers $A$ and $B$, we used a paired bootstrap: the same resampled indices were applied to both systems so that trial-level correlation was retained. For each resample, we recorded the metric difference and its sign,
\begin{equation}
\begin{aligned}
\Delta_m^{(b)} &= m(A^{(b)}) - m(B^{(b)}), \\
p_{A>B} &= \frac{1}{B}\sum_{b} \mathbf{1}\!\left[\Delta_m^{(b)} > 0\right].
\end{aligned}
\end{equation}
A difference was treated as significant when the interval for $\Delta_m$ excluded zero. The main result is a sign change in which different systems are identified as significantly better by full AUC and by the strict-threshold metric. The classical comparison for two full areas is the method of DeLong~\cite{delong1988comparing}; here, the bootstrap extends the significance statement to the low-FMR region as well. Resampling was performed at the trial-pair level. Because the same identities appear in many pairs, a pair-level bootstrap can understate variance, so an identity-clustered bootstrap is the more conservative option. This point is noted as a limitation and should be checked in future work, especially for the face ranking flip. The analysis used common scientific Python tools~\cite{pedregosa2011scikit}.

\section{Results and Discussion}


\subsection{Score distributions and operating thresholds}
\begin{figure*}[tp]
\centering
\includegraphics[width=\textwidth]{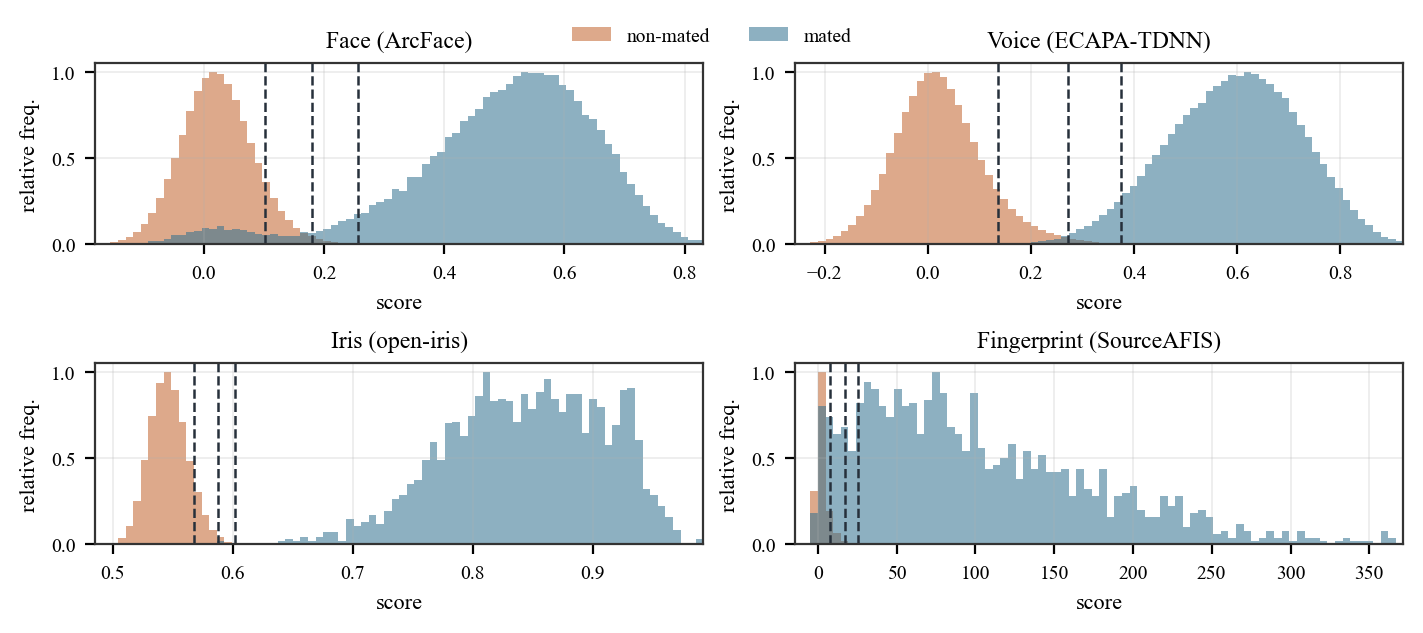}
\caption{Mated and non-mated score distributions per modality, with the fixed FMR thresholds at $10^{-1}$, $10^{-2}$, and $10^{-3}$ marked. The operating thresholds lie in the non-mated tail, where the mated acceptance changes quickly.}
\label{fig:scores}
\end{figure*}

\begin{figure*}[tp]
\centering
\includegraphics[width=\textwidth]{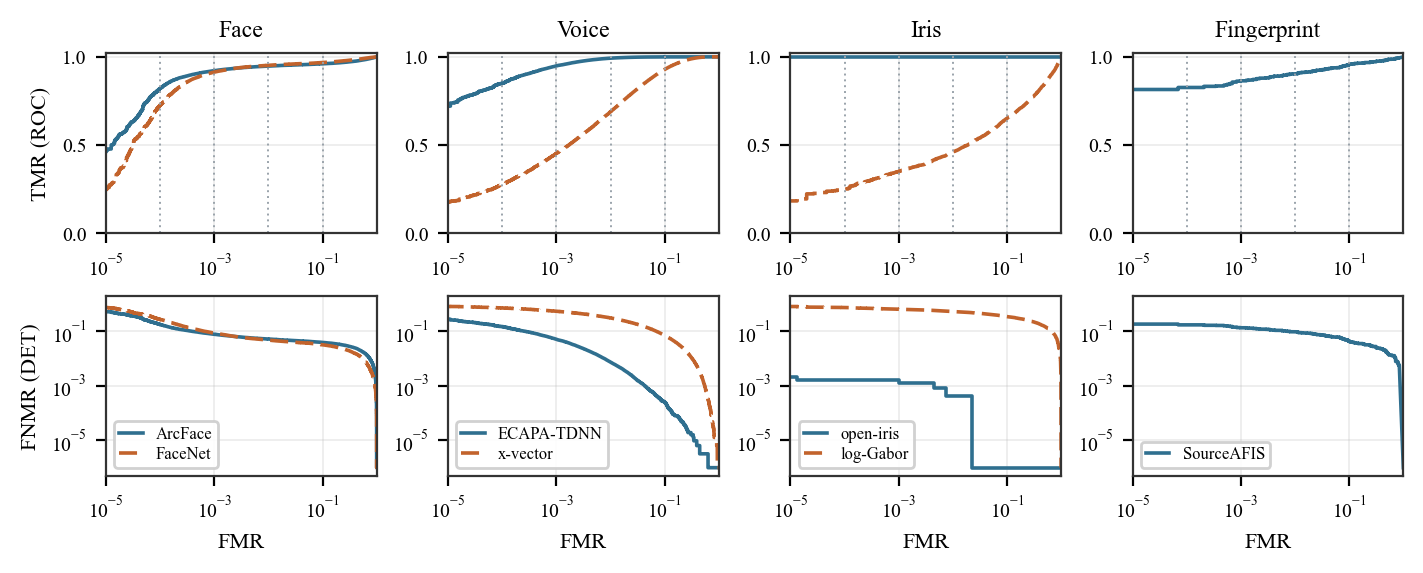}
\caption{ROC curves on a logarithmic FMR axis, top row, and DET curves, bottom row, with the two matchers of each modality overlaid. The DET curve is the primary reporting view; a crossing of the face matchers is observed in the strict region, from which the ranking flip is produced.}
\label{fig:roc}
\end{figure*}

Figure~\ref{fig:scores} shows the mated and non-mated score distributions for all four modalities, with fixed-FMR thresholds marked. The strict thresholds lie deep in the non-mated tail, where a small threshold movement can noticeably change mated acceptance. This tail behavior explains why a full-curve summary can miss the part of the score distribution that determines the deployed decision. Figure~\ref{fig:roc} shows the same systems as ROC curves on a logarithmic FMR axis and as DET curves, making the low-FMR region visible instead of compressing it into a corner.

\begin{figure*}[tp]
\centering
\includegraphics[width=\textwidth]{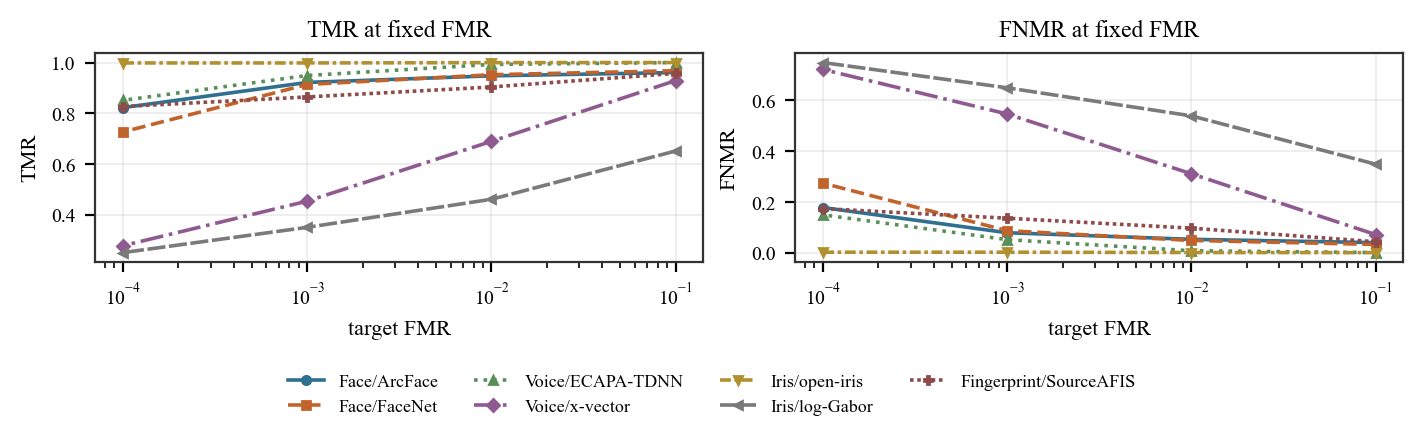}
\caption{TMR and FNMR at fixed FMR targets. Deployment-relevant performance is represented by the operating points, which become difficult to resolve for the small datasets as the target FMR is decreased.}
\label{fig:fixed}
\end{figure*}

\subsection{Overstatement of strict-threshold performance}
\begin{figure}[tbp]
\centering
\includegraphics[width=\linewidth]{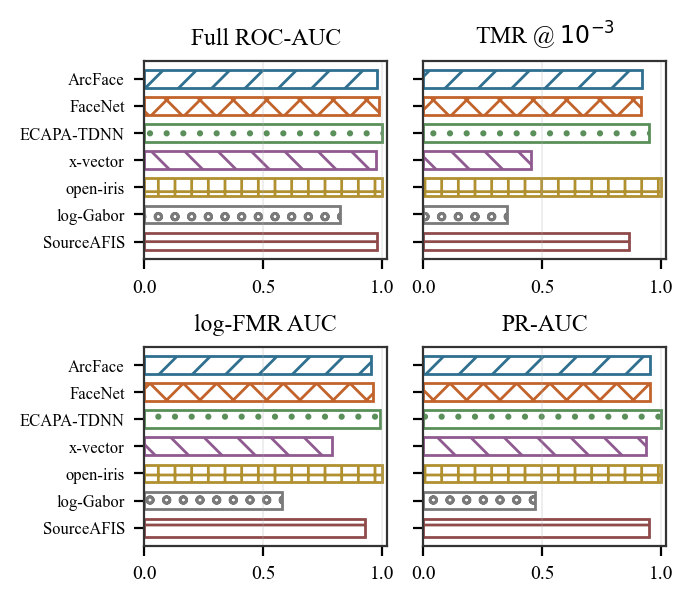}
\caption{Four of the metrics across all seven systems; the remaining metrics are listed in Table~\ref{tab:main}. The systems are compressed into a narrow high band by the full ROC-AUC, whereas greater separation and the weaker systems are revealed by the strict-threshold metrics.}
\label{fig:permetric}
\end{figure}

Table~\ref{tab:main} reports the metrics for all seven systems, including $95$ percent intervals for the full AUC and for TMR at $\mathrm{FMR}=10^{-3}$. Figure~\ref{fig:permetric} plots four of the metrics across systems. Across the experiments, the full AUC was consistently higher than the strict-threshold value. The clearest example is the x-vector speaker verifier: it reached a full AUC of $0.975$, but its TMR at $\mathrm{FMR}=10^{-3}$ was $0.453$. By contrast, ECAPA-TDNN kept a TMR of $0.949$ at the same threshold. The full AUC gap between these two voice systems was only $0.024$, whereas the TMR gap at $10^{-3}$ was $0.495$. Thus, the full AUC understated the operating-region difference by roughly a factor of twenty. Similar reductions appear for the classical iris encoder and the fingerprint matcher. Overall, full ROC-AUC compresses the systems into a narrow high band, while the strict-threshold metrics reveal much larger differences, especially for weaker matchers. A high full AUC therefore should not be read as evidence of strong low-FMR performance unless the operating-point result is also shown.

\begin{table*}[t]
\centering
\caption{Per-system metrics. Full ROC-AUC and the true match rate at $\mathrm{FMR}=10^{-3}$ are shown with $95\%$ bootstrap confidence intervals; the recommended operating-point report is the complementary $\mathrm{FNMR}=1-\mathrm{TMR}$. minDCF is lower is better, all other columns are higher is better.}
\label{tab:main}
\footnotesize
\begin{tabular}{@{}llcccccccc@{}}
\toprule
Modality & Matcher & Full AUC [95\% CI] & TMR@$10^{-3}$ [95\% CI] & EER & log-FMR AUC & PR-AUC & minDCF & $d'$ \\
\midrule
Face & ArcFace & 0.979 [0.978, 0.980] & 0.922 [0.919, 0.924] & 0.044 & 0.955 & 0.952 & 0.139 & 3.84 \\
Face & FaceNet & 0.985 [0.985, 0.986] & 0.914 [0.911, 0.916] & 0.039 & 0.960 & 0.955 & 0.164 & 4.31 \\
Voice & ECAPA-TDNN & 1.000 [1.000, 1.000] & 0.949 [0.946, 0.951] & 0.009 & 0.991 & 0.999 & 0.123 & 5.17 \\
Voice & x-vector & 0.975 [0.975, 0.976] & 0.453 [0.448, 0.459] & 0.084 & 0.788 & 0.939 & 0.645 & 2.45 \\
Iris & open-iris & 1.000 [1.000, 1.000] & 0.998 [0.997, 0.999] & 0.001 & 1.000 & 1.000 & 0.002 & 6.13 \\
Iris & log-Gabor & 0.825 [0.819, 0.831] & 0.350 [0.334, 0.366] & 0.251 & 0.580 & 0.468 & 0.725 & 1.16 \\
Fingerprint & SourceAFIS & 0.978 [0.975, 0.980] & 0.864 [0.852, 0.876] & 0.061 & 0.929 & 0.949 & 0.180 & 1.87 \\
\bottomrule
\end{tabular}
\end{table*}

\subsection{Metric-dependent system ranking}
The strongest example of metric-dependent ranking appears in the face results, where the two matchers are close enough that the preferred system depends on the metric being reported. Table~\ref{tab:flip} gives the paired bootstrap difference, ArcFace minus FaceNet. For full AUC, the difference is negative and its interval excludes zero, so FaceNet has the higher full-curve summary. For TMR at $\mathrm{FMR}=10^{-3}$, low-FMR partial AUC, and minDCF, the sign reverses and the intervals again exclude zero, so ArcFace performs better in the strict operating region. Across bootstrap resamples, the estimated probability that ArcFace exceeded FaceNet was $0.00$ for full AUC and $1.00$ for TMR at $10^{-3}$. A report led by full AUC would therefore select FaceNet, while a report led by the operating point would select ArcFace. No ranking flip was found for voice or iris, because the stronger system remained better at every operating point. Even there, however, the apparent size of the difference depended strongly on the metric. Metric choice is therefore not only a matter of presentation; it can change the conclusion of a system comparison. For that reason, a fair comparison should include a paired test at the deployment threshold rather than rely on AUC ordering alone.

\begin{table}[t]
\centering
\caption{Face, paired bootstrap difference (ArcFace $-$ FaceNet) with the $95\%$ interval. The system identified as significantly better is reversed between the full AUC and the strict-threshold metrics. Lower minDCF is better, so ArcFace is favoured by a negative difference.}
\label{tab:flip}
\footnotesize
\begin{tabular}{@{}lccc@{}}
\toprule
Metric & $\Delta$ mean & $95\%$ CI & Better system \\
\midrule
Full ROC-AUC & $-0.0065$ & $[-0.0071, -0.0058]$ & FaceNet \\
TMR@$10^{-3}$ & $+0.0082$ & $[+0.0057, +0.0111]$ & ArcFace \\
low-FMR pAUC & $+0.0019$ & $[+0.0005, +0.0034]$ & ArcFace \\
minDCF & $-0.0248$ & $[-0.0290, -0.0213]$ & ArcFace \\
\bottomrule
\end{tabular}
\end{table}

\subsection{Optimism under class imbalance}
For every system, PR-AUC was lower than ROC-AUC, and the gap widened as performance weakened, as shown in Table~\ref{tab:main}. The classical iris encoder provides the largest example: its ROC-AUC was $0.825$, while its PR-AUC was $0.468$, a gap of $0.357$. The x-vector showed a smaller but still visible gap of $0.037$, whereas the strong ECAPA-TDNN system showed a gap of only $0.001$. Because the non-mated class is much larger than the mated class, a system can keep a high ROC-AUC while producing poor precision at relevant thresholds. The PR-AUC gap therefore offers a second indication that ROC-AUC alone is not enough for reporting verification performance under imbalance~\cite{davis2006prroc}. The same imbalance that makes low-FMR operation difficult also makes a global ROC scalar look overly optimistic.

\subsection{Uncertainty and measurement resolution}
Bootstrap intervals are included for full AUC and strict-threshold TMR in Table~\ref{tab:main}, and for the paired differences in Table~\ref{tab:flip}. These intervals show that the face ranking flip is not simply a small numerical fluctuation. They also reveal a practical limit: the smallest FMR that a test can resolve is roughly the reciprocal of the number of non-mated trials. A claim at $\mathrm{FMR}=10^{-4}$ is therefore not supported by the fingerprint set with about $14{,}000$ within-sensor non-mated pairs in the same way that it is supported by the face set with two million pairs. Figure~\ref{fig:fixed} shows the fixed-FMR operating points, where the smaller datasets become less stable as the target FMR decreases. A low-FMR value should therefore be reported with its interval and trial count; otherwise, the reader cannot judge its measurement resolution. This is the same uncertainty requirement emphasized by the standard~\cite{efron1993bootstrap,grother2019frvt}.

\begin{table}[t]
\centering
\caption{A short reporting checklist for biometric verification, aligned with ISO/IEC 19795-1. With each item, a source of misinterpretation from a single full AUC is removed.}
\label{tab:checklist}
\footnotesize
\begin{tabular}{@{}p{2.55cm}p{4.9cm}@{}}
\toprule
Item & What to report \\
\midrule
Primary graphic & DET curve, FNMR against FMR, with the low-FMR region legible \\
Operating points & FNMR at the deployment FMR ($10^{-2}$ to $10^{-4}$) \\
Uncertainty & A confidence interval and the trial count on every value \\
Trial protocol & Mated and non-mated construction, score direction, and the number of non-mated trials \\
Test conditions & Dataset, population, and sensor or protocol \\
Supplementary scalars & ROC-AUC and EER, with the AUC interval and integration variable stated \\
Imbalance & PR-AUC alongside ROC-AUC \\
\bottomrule
\end{tabular}
\end{table}

\section{Conclusion and Future Work}

Full ROC-AUC remains a useful global summary of ranking performance, but it does not by itself describe the low-FMR region where many biometric verifiers are deployed. In this study of seven pretrained systems across face, voice, iris, and fingerprint, strict-threshold behavior was often much lower than the full AUC suggested. In the face comparison, the system identified as better changed with the headline metric: FaceNet had the higher full AUC, while ArcFace had the higher TMR at $\mathrm{FMR}=10^{-3}$ and the stronger strict-threshold summaries. These findings support the reporting style already encouraged by ISO/IEC 19795-1. The DET curve and FNMR at a fixed FMR should be the primary report, with low-FMR partial AUC or log-FMR AUC and minimum detection cost used as supporting summaries. Each value should include a confidence interval and the relevant trial count. Full ROC-AUC and EER can still be useful, but they should appear as context rather than as the main evidence for deployment performance. Systems should also be compared with a paired test at the threshold of interest, not only by AUC ordering. The checklist in Table~\ref{tab:checklist} summarizes these reporting items. Future work should add an identity-clustered bootstrap, subject-disjoint and cross-sensor protocols, calibration-aware costs across several priors, and demographic stratification, so that operating-region reports remain reliable across both thresholds and populations.

\bibliographystyle{IEEEtran}
\bibliography{egbib}

\end{document}